УДК 004.93

*Коробейников А.В.,*

*к.т.н., директор, ООО «ИжТелеМед», г. Ижевск;*

*Исламгалиев И.И.,*

*инженер-программист, ОАО «Сарапульский электрогенераторный завод»*

# МОДИФИКАЦИЯ АЛГОРИТМА КОНЦЕПТУАЛЬНОЙ КЛАСТЕРИ-ЗАЦИИ COBWEB ДЛЯ КОЛИЧЕСТВЕННЫХ ДАННЫХ С ИПОЛЬ-ЗОВАНИЕМ НЕЧЕТКОЙ ФУНКЦИИ ПРИНАДЛЕЖНОСТИ

**Аннотация.** Предлагается модификация алгоритма концептуальной кластеризации *Cobweb* с целью применения его для количественных данных.

**Ключевые слова:** кластеризация, алгоритм *Cobweb*, количественные данные, нечеткая функция принадлежности.

**Korobeynikov A.V.,** cand.tech.sci., director, Ltd «IzhTeleMed»;

**Islamgaliev I.I.,** software engineer, JSC «Sarapul Electric Generators»

# MODIFICATION OF CONCEPTUAL CLUSTERING ALGORITHM COBWEB FOR NUMERICAL DATA USING FUZZY MEMBERSHIP FUNCTION

**Abstract.** Modification of a conceptual clustering algorithm Cobweb for the purpose of its application for numerical data is offered.

***Алгоритм концептуальной кластеризации Cobweb.*** В задачах кластеризации одной из проблем является обоснование количества необходимых категорий (кластеров). В алгоритме *Cobweb* для корректного определения количества кластеров, глубины иерархии и принадлежности категории новых экземпляров используются эвристика – глобальная метрика качества. В алгоритме *Cobweb* реализован инкрементальный алгоритм обучения, не требующий представления всех обучающих примеров до начала обучения.

Алгоритм осуществляет разделение на кластеры основываясь на понятии полезности разбиения (*category utility*) [1]:

$$CU = \sum_{k=1}^{n} \sum_{j} \sum_{i} P(A_j = v_{ij})P(C_k \mid A_j = v_{ij})P(A_j = v_{ij} \mid C_k); \quad (1)$$

где $C_k$ – $k$-ый кластер, $n$ – количество кластеров, $A_j$ – $j$-ый параметр образца, $v_{ij}$ – $i$-ое значение $j$-го параметра. Значение $P(A_j = v_{ij} \mid C_k)$ называется предсказуемостью (*predictability*). Это вероятность того, что у образца свойство $A_i$ принимает значение $v_{ij}$, при условии, что он относится к категории $C_k$. Величина $P(C_k \mid A_j = v_{ij})$ называется предиктивностью (*predictiveness*). Это вероятность того, что образец относится к категории $C_k$, при условии, что свойство $A_j$ принимает значение $v_{ij}$. Значение $P(A_j = v_{ij})$ – это весовой коэффициент, усиливающий влияние наиболее распространенных свойств. Благодаря совместному учету этих значений высокая полезность разбиения на категории (*CU*) означает высокую вероятность того, что объекты из одной категории обладают одинаковыми свойствами, и низкую вероятность наличия этих свойств у объектов из других категорий.

Критерий полезности категории был определен при исследовании человеческой категоризации. Он учитывает влияние категорий базового уровня и другие аспекты структуры человеческих категорий [1].

При предъявлении нового образца алгоритм *Cobweb* оценивает полезность разбиений при отнесении образца к одной из существующих категорий, а также полезность возможных модификаций иерархии категорий.

В начале работы алгоритма вводится единственная категория, свойства которой совпадают со свойствами первого образца. Для каждого очередного образца алгоритм начинает свою работу с корневой категории и движется далее по дереву. На каждом уровне выполняется оценка эффективности категоризации на основе полезности разбиения. При этом оцениваются результаты следующих модификаций дерева категорий [1]:

1) отнесение образца к наилучшей из существующих категорий;

2) добавление новой категории, содержащей единственный образец;

3) слияние двух категорий в одну с добавлением текущего образца;

4) разбиение существующей категории на две и отнесение текущего образца к лучшей из вновь созданных категорий.

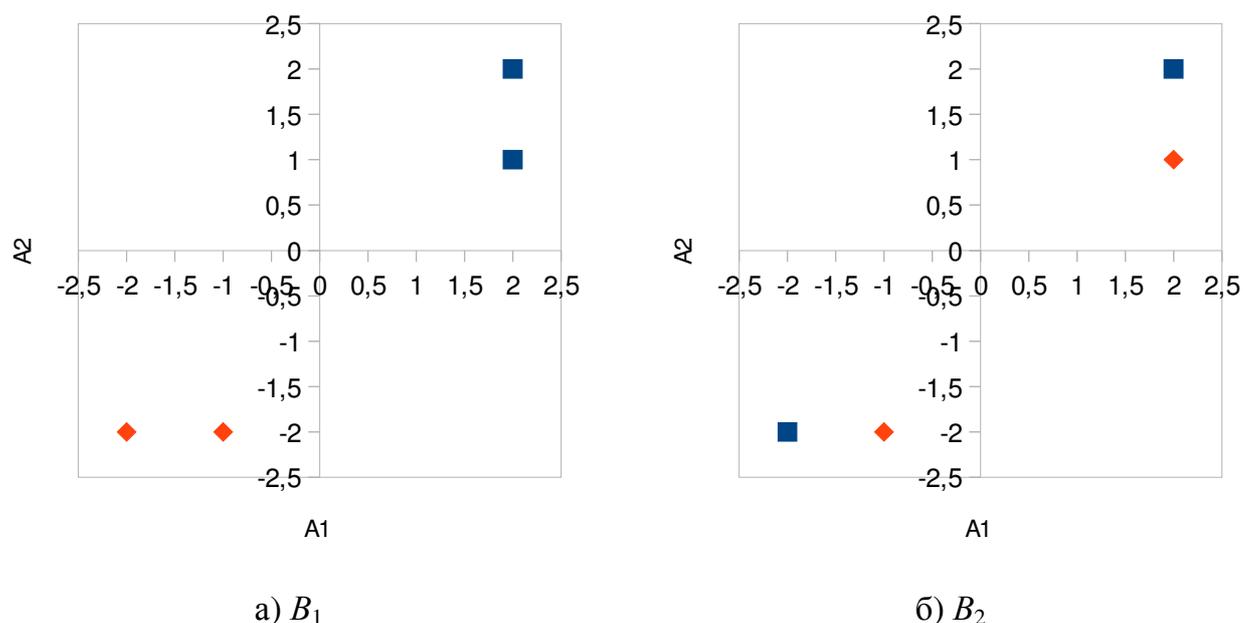

а) $B_1$ б) $B_2$

*Рисунок 1.* Варианты разбиения образцов с равными параметрами

*Таблица 1*

| $k$ | $j$ | $v_{ij}$ | $P(A_j=v_{ij})$ | | $P(A_j=v_{ij}|C_k)$ | | $P(C_k|A_j=v_{ij})$ | | $CU$ | |
|---|---|---|---|---|---|---|---|---|---|---|
| | | | $B_1$ | $B_2$ | $B_1$ | $B_2$ | $B_1$ | $B_2$ | $B_1$ | $B_2$ |
| $C_1$ | $A_1$ | -2 | 1/4 | 1/4 | 0 | 1/2 | 0 | 1 | 0 | 1/8 |
| | | -1 | 1/4 | 1/4 | 0 | 0 | 0 | 0 | 0 | 0 |
| | | 1 | 0 | 0 | 0 | 0 | 0 | 0 | 0 | 0 |
| | | 2 | 1/2 | 1/2 | 1 | 1/2 | 1 | 1/2 | 1/2 | 1/8 |
| | $A_2$ | -2 | 1/2 | 1/2 | 0 | 1/2 | 0 | 1/2 | 0 | 1/8 |
| | | -1 | 0 | 0 | 0 | 0 | 0 | 0 | 0 | 0 |
| | | 1 | 1/4 | 1/4 | 1/2 | 0 | 1 | 0 | 1/8 | 0 |
| | | 2 | 1/4 | 1/4 | 1/2 | 1/2 | 1 | 1 | 1/8 | 1/8 |
| $C_2$ | $A_1$ | -2 | 1/4 | 1/4 | 1/2 | 0 | 1 | 1 | 1/8 | 0 |
| | | -1 | 1/4 | 1/4 | 1/2 | 1/2 | 1 | 0 | 1/8 | 0 |
| | | 1 | 0 | 0 | 0 | 0 | 0 | 0 | 0 | 0 |
| | | 2 | 1/2 | 1/2 | 0 | 1/2 | 0 | 1/2 | 0 | 1/8 |
| | $A_2$ | -2 | 1/2 | 1/2 | 1 | 1/2 | 1 | 1/2 | 1/2 | 1/8 |
| | | -1 | 0 | 0 | 0 | 0 | 0 | 0 | 0 | 0 |
| | | 1 | 1/4 | 1/4 | 0 | 1/2 | 0 | 1 | 0 | 1/8 |
| | | 2 | 1/4 | 1/4 | 0 | 0 | 0 | 0 | 0 | 0 |
| | | | | | | | | | **1,500** | **0,875** |

Алгоритм эффективен и выполняет кластеризацию на разумное число классов. В нем используется вероятностное представление принад-лежности, и получаемые категории являются гибкими и робастными [1].

Рассмотрим работу алгоритма на примере. На рис. 1 представлены варианты разбиения. Результаты работы алгоритма, приведенные в табл. 1, показывают предпочтительность разбиения $B_1$ перед $B_2$.

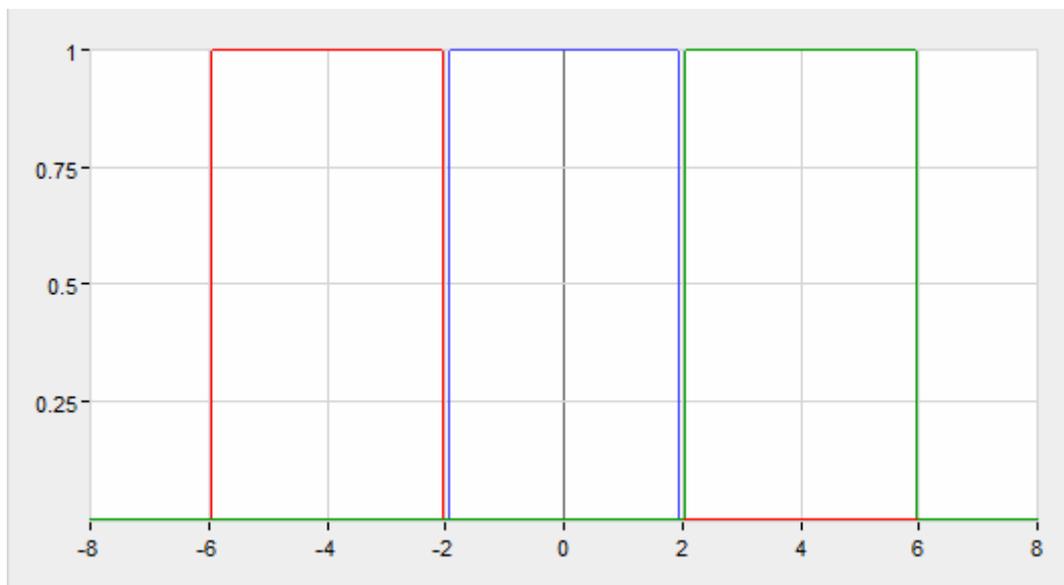

а) прямоугольная

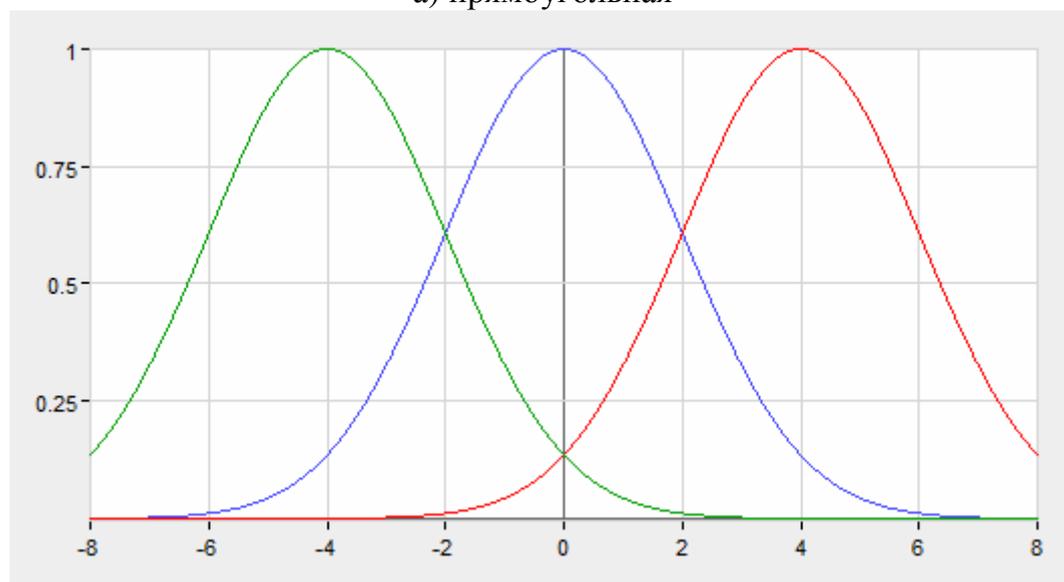

б) на основе нормального распределения

*Рисунок 2.* Функции принадлежности отсчетов

Недостатком данного варианта алгоритма является возможность работы только с качественными данными (перечислимыми значениями), например, перечень цветов: желтый, красный, синий. Если нет совпадений значений параметров образцов, то близость или удаленность образцов на координатной плоскости параметров никак не влияет на результат.

***Алгоритм кластеризации Cobweb для количественных данных.***
Для работы с количественными данными (диапазон значений) возможна модификация алгоритма: передискретизация значений на числовых осях параметров (рис. 2а, прямоугольная функция принадлежности), то есть сведение количественных данных к качественным. Однако при таком подходе значения параметров, находящиеся на числовой оси параметров рядом, но на границе функций принадлежности, могут попасть в разные отсчеты ($v_{ij}$).

В данной работе для устранения недостатков описанных вариантов алгоритма *Cobweb* предлагается использовать нечеткую функцию принадлежности на основе формулы нормального распределения [2] (рис. 2б):

$$f(m,i,j) = \exp\left(-\frac{(a_{mj} - v_{ij})^2}{2\sigma_j^2}\right); \qquad (2)$$

где $a_{mj}$ – $m$-ное значение параметра $A_j$ (случайной величины); $v_{ij}$ – $i$-е значение центра отсчета параметра $A_j$ (мат. ожидание); $\sigma_j^2$ – дисперсия $A_j$.

Для замены вероятностей предлагается использовать формулы:

$$\overline{P}(A_j = v_{ij} \mid C_k) = \frac{1}{l_k}\sum_{m=1}^{l_k} f(m,i,j) \qquad (3)$$

$$\overline{P}(C_k \mid A_j = v_{ij}) = \frac{\sum_{m=1}^{l_k} f(m,i,j)}{\sum_{m=1}^{l} f(m,i,j)} \qquad (4)$$

$$\overline{P}(A_j = v_{ij}) = \frac{1}{l}\sum_{m=1}^{l} f(m,i,j) \qquad (5)$$

где $l_k$ – количество элементов в кластере $C_k$; $l$ – общее количество элементов. Эвристика полезности разбиения с учетом изменений:

$$\overline{CU} = \sum_{k=1}^{n}\sum_{j=1}^{q}\sum_{i=1}^{d} \overline{P}(A_j = v_{ij})\overline{P}(C_k \mid A_j = v_{ij})\overline{P}(A_j = v_{ij} \mid C_k) \qquad (6)$$

где $n$-количество элементов в кластере, $q$ – количество свойств, $d$ – количество отсчетов параметра $A_j$.

Значения $v_{ij}$ представляют собой узлы некоторой сетки размером $d$:

$$v_{ij} = (i - 0.5)\frac{\max(A_j) - \min(A_j)}{d}; i = 1..d \qquad (7)$$

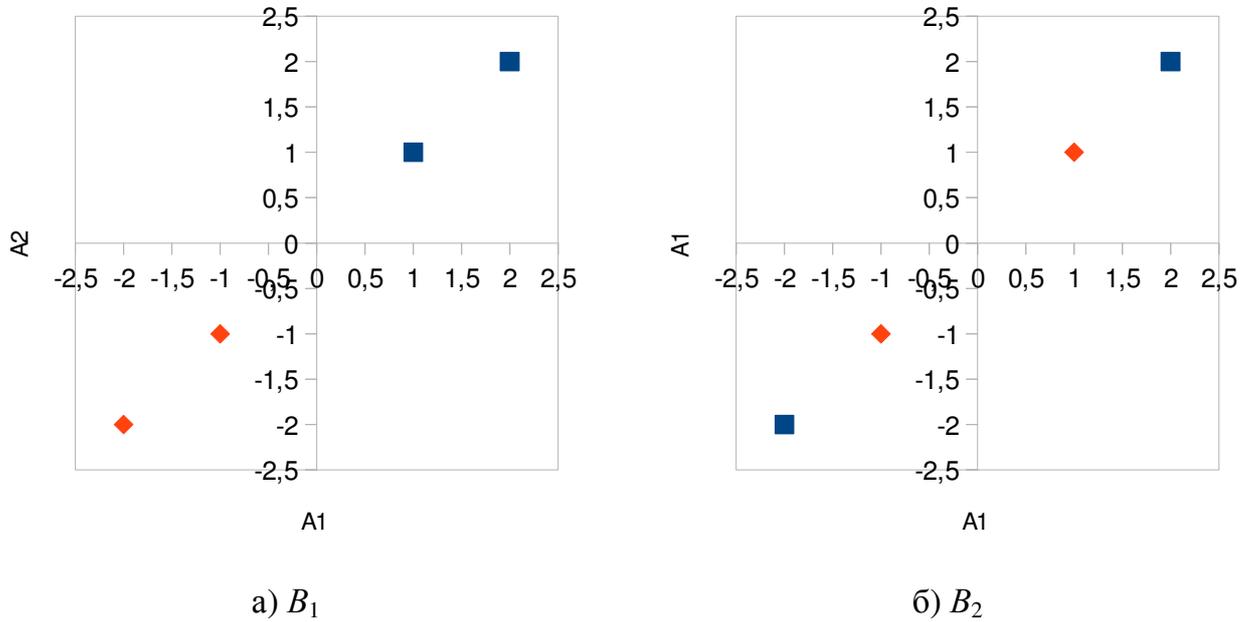

а) $B_1$           б) $B_2$

*Рисунок 3.* Варианты разбиения образцов с неравными параметрами

*Таблица 2*

| $k$ | $j$ | $v_{ij}$ | $\overline{P}(A_j = v_{ij})$ | | $\overline{P}(A_j = v_{ij} \mid C_k)$ | | $\overline{P}(C_k \mid A_j = v_{ij})$ | | $\overline{CU}$ | |
|---|---|---|---|---|---|---|---|---|---|---|
| | | | $B_1$ | $B_2$ | $B_1$ | $B_2$ | $B_1$ | $B_2$ | $B_1$ | $B_2$ |
| $C_1$ | $A_1$ | -2 | 0,006 | 0,500 | 0,007 | 0,618 | 0,404 | 0,404 | 0,000 | 0,125 |
| | | -1 | 0,006 | 0,500 | 0,007 | 0,618 | 0,404 | 0,404 | 0,000 | 0,125 |
| | | 1 | 0,073 | 0,309 | 0,084 | 0,352 | 0,438 | 0,438 | 0,003 | 0,048 |
| | | 2 | 0,073 | 0,309 | 0,084 | 0,352 | 0,438 | 0,438 | 0,003 | 0,048 |
| | $A_2$ | -2 | 0,803 | 0,309 | 0,916 | 0,352 | 0,438 | 0,438 | 0,323 | 0,048 |
| | | -1 | 0,803 | 0,309 | 0,916 | 0,352 | 0,438 | 0,438 | 0,323 | 0,048 |
| | | 1 | 0,803 | 0,500 | 0,993 | 0,618 | 0,404 | 0,404 | 0,323 | 0,125 |
| | | 2 | 0,803 | 0,500 | 0,993 | 0,618 | 0,404 | 0,404 | 0,323 | 0,125 |
| $C_2$ | $A_1$ | -2 | 0,803 | 0,309 | 0,993 | 0,382 | 0,404 | 0,404 | 0,323 | 0,048 |
| | | -1 | 0,803 | 0,309 | 0,993 | 0,382 | 0,404 | 0,404 | 0,323 | 0,048 |
| | | 1 | 0,803 | 0,568 | 0,916 | 0,648 | 0,438 | 0,438 | 0,323 | 0,161 |
| | | 2 | 0,803 | 0,568 | 0,916 | 0,648 | 0,438 | 0,438 | 0,323 | 0,161 |
| | $A_2$ | -2 | 0,073 | 0,568 | 0,084 | 0,648 | 0,438 | 0,438 | 0,003 | 0,161 |
| | | -1 | 0,073 | 0,568 | 0,084 | 0,648 | 0,438 | 0,438 | 0,003 | 0,161 |
| | | 1 | 0,006 | 0,309 | 0,007 | 0,382 | 0,404 | 0,404 | 0,000 | 0,048 |
| | | 2 | 0,006 | 0,309 | 0,007 | 0,382 | 0,404 | 0,404 | 0,000 | 0,048 |
| | | | | | | | | | **2,592** | **1,526** |

Рассмотрим пример работы модифицированного алгоритма (на рис. 3). Результаты работы алгоритма, при размере сетки (количестве отсчетов) $d = 4$ и $\sigma_j = 1$ для данного примера приведены в табл. 2.

На рис. 4 представлена экранная форма программного обеспечения, которое было разработано для проверки предложенного метода.

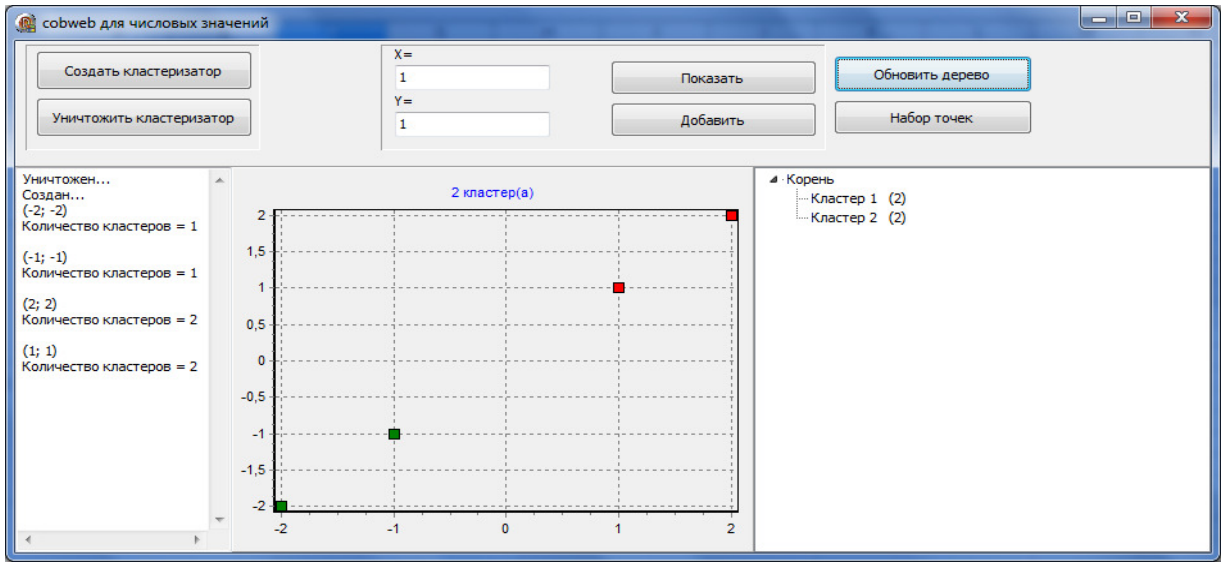

*Рисунок 4.* Экранная форма программы

**Список литературы**

1. Д. Ф. Люгер. Искусственный интеллект: стратегии и методы решения сложных проблем. – М. Издательский дом «Вильямс», 2003.

2. Гурман В.Е. Теория вероятностей и математическая статистика: Учебное пособие для вузов. – 9-е изд. – М.: Высш. шк., 2003. – 479с.